\documentclass[letterpaper,journal]{IEEEtran}
\usepackage{amsmath,amsfonts}
\usepackage{algorithm}
\usepackage{algorithmic}
\usepackage{array}
\usepackage{booktabs}
\usepackage{multirow}
\usepackage[caption=false,font=normalsize,labelfont=sf,textfont=sf]{subfig}
\usepackage{textcomp}
\usepackage{stfloats}
\usepackage{url}
\usepackage{verbatim}
\usepackage{graphicx}
\usepackage{cite}
\makeatletter
\renewcommand\section{\@startsection{section}{1}{\z@}%
  {1.3ex plus 0.1ex minus 0.2ex}%
  {0.35ex plus 0.1ex minus 0.1ex}%
  {\normalfont\normalsize\centering\scshape}}
\renewcommand\subsection{\@startsection{subsection}{2}{\z@}%
  {1.3ex plus 0.3ex minus 0.3ex}%
  {0.3ex plus 0.2ex minus 0.1ex}%
  {\normalfont\normalsize\itshape}}
\makeatother
\def\BibTeX{{\rm B\kern-.05em{\sc i\kern-.025em b}\kern-.08em
    T\kern-.1667em\lower.7ex\hbox{E}\kern-.125emX}}
\usepackage{balance}
\usepackage[hidelinks]{hyperref}
\begin{document}
\bstctlcite{IEEEexample:BSTcontrol}
\title{Horizon-Coupled Obstacle Guidance for Relative-Action Diffusion Policies}
\author{Guangyuan Wang\textsuperscript{*,1},
Haishan Xia\textsuperscript{*,1}, Zitao Zhang\textsuperscript{*,1},
Kejia Chen\textsuperscript{1}, Yuanchen Sun\textsuperscript{1},
Zhenguo Sun\textsuperscript{1}, Donghao Zhang\textsuperscript{2},
Yu Zhang\textsuperscript{\textdagger,1}, and Alois Knoll\textsuperscript{1}%
\thanks{\textsuperscript{*}These authors contributed equally.
\textsuperscript{\textdagger}Yu Zhang is the corresponding author.}%
\thanks{\textsuperscript{1}School of Computation, Information and Technology,
Technical University of Munich, Germany.}%
\thanks{\textsuperscript{2}Beijing Information Science and Technology University,
China.}%
\thanks{This work has been submitted to the IEEE for possible publication.
Copyright may be transferred without notice, after which this version may no
longer be accessible.}}

\markboth{IEEE Robotics and Automation Letters}%
{How to Use the IEEEtran \LaTeX \ Templates}

\maketitle

\begin{abstract}

Real-world robot deployment requires adapting learned policies to unseen obstacles without collecting 
new demonstrations or retraining. Existing policy-level guidance often applies obstacle corrections independently at individual horizon steps, 
which may address local obstacle violations but fail to coordinate the action sequence into a complete, task-directed detour.
How obstacle responses should be propagated across the 
action horizon to coordinate complete, task-directed avoidance motion remains underexplored. 
We introduce an inference-time obstacle-guidance framework for pretrained relative-action diffusion 
policies. It rolls out cumulative relative actions into a workspace trajectory, evaluates obstacle 
interactions along the trajectory with an SDF-based obstacle model, and propagates each obstacle response to all contributing 
action increments during denoising. Real-robot experiments across seven obstacle-constrained scenarios spanning
four manipulation tasks show that our method achieves 105/140 (75.0\%)
collision-free task successes, compared with 8/140 (5.7\%) for
absolute-action point-wise guidance. Ablations show that the relative-action horizon-coupled formulation provides the 
larger observed improvement across both obstacle models.
\end{abstract}
\vspace{-1.34ex}
\vspace{-0.67ex}

\section{Introduction}

Diffusion Policy (DP)~\cite{chi2023diffusionpolicy} has emerged as a prominent
framework for visuomotor manipulation, owing to its ability to model multimodal
action distributions from expert demonstrations and generate observation-conditioned action chunks for
closed-loop control. Recent 3D variants~\cite{Ze2024DP3,ke2025threediffuseractor,
ryu2024diffusionedfs,yang2025equibot} incorporate explicit 3D scene
representations to improve spatial generalization across complex manipulation
tasks. These advances strengthen the potential of DPs for deployment in complex manipulation environments.
However, real-world deployment may introduce obstacles that are absent from the expert demonstrations utilized to train 
the policy, and these advances do not by themselves enable the pretrained policy to adapt its motion around 
them. Such obstacles can block the learned motion even when 
the manipulation objective remains unchanged. A straightforward solution 
is to redesign the robot motion for each obstacle configuration, but repeated 
redesign requires substantial engineering time, human effort, and deployment cost. 
Adapting the sampled actions of a trained DP to deployment-time obstacles without 
retraining would avoid these repeated costs and make policy deployment more scalable.

A practical way to address this deployment challenge is to adapt the actions
sampled by the trained DP to newly introduced obstacles. Inference-time policy
steering~\cite{wang2025itps} provides this mechanism without retraining the base
policy. One route learns an auxiliary steering policy over diffusion noise through 
reinforcement learning~\cite{wagenmaker2025dsrl}, requiring reward signals and 
additional interaction for policy optimization. In the reported real-world setting, 
this interaction is collected through hardware rollouts, which may involve unsafe actions 
and potential collisions. Such interaction makes reinforcement-learning-based steering 
less suitable for safety-critical deployment-time adaptation. A second route modifies reverse
diffusion with gradients derived from externally specified robot
objectives~\cite{lee2025kcgg,du2025dynaguide,gupta2026umionair}. Obstacle
geometry directly defines such a differentiable objective, eliminating
auxiliary-policy training and reward-bearing exploration. We therefore adopt
denoising guidance.

Existing denoising-guidance methods demonstrate that safety constraints can
be incorporated into reverse sampling. Planner-centric methods such as
MPD~\cite{carvalho2025mpd}, SafeDiffuser~\cite{xiao2025safediffuser},
DPCC~\cite{romer2025dpcc}, and Constrained
Diffusers~\cite{zhang2025constrained} formulate obstacle avoidance as
constrained state- or joint-trajectory generation under specified state or
goal conditions. Such formulations address state-to-goal motion planning
rather than adapting the action generation of a pretrained visuomotor policy
from evolving visual feedback. Extending obstacle avoidance from this
planner-centric setting to learned manipulation under visual feedback therefore
requires guidance at the policy level.
Lan-o3dp~\cite{11127231} and EmbodiSteer~\cite{wang2026embodisteer}
guide action chunks from pretrained visuomotor policies. Both apply obstacle corrections independently at individual 
horizon steps, without explicitly coordinating avoidance responses across the action horizon. Such step-wise corrections 
may address local obstacle violations but fail to organize the action sequence into a complete detour that both avoids 
the obstacle and continues toward the task goal. How obstacle responses should be propagated across the 
action horizon to coordinate complete, task-directed avoidance motion remains underexplored.

To close this gap, we introduce horizon-coupled obstacle guidance for 
pretrained relative-action DPs in static environments. Relative end-effector targets are
formed by recursively integrating positional increments from the preceding
target~\cite{aljalbout2024actionspace}. From this recurrence, we derive a
horizon-coupled guidance structure in which each predicted position depends on all
preceding increments. At each denoising step, our method rolls out the
estimated action chunk into a predicted end-effector trajectory in the
workspace for obstacle evaluation and propagates each response to every
increment contributing to the affected position. This horizon-coupled propagation 
coordinates actions across the horizon to form the avoidance motion. The frozen DP supplies the learned
task behavior throughout guided sampling.

In summary, our main contributions are:
(1) a horizon-coupled obstacle-guidance formulation for relative-action diffusion policies that propagates obstacle responses across the action horizon, enabling coordinated, task-directed avoidance motion rather than isolated local corrections;
(2) an inference-time guided-sampling framework that applies this cross-step propagation to predicted end-effector trajectories during denoising, allowing deployment-time obstacle constraints to reshape the action sequence without policy retraining;
(3) real-robot validation across seven obstacle-constrained scenarios spanning four manipulation tasks, achieving 105/140 (75.0\%) collision-free task successes compared with 8/140 (5.7\%) for absolute-action point-wise guidance.

\section{Preliminaries and Problem Formulation}
\label{sec:preliminaries}
\noindent In this section, we first define the visuomotor control setting, then
review diffusion-based obstacle guidance and the limitations of existing
methods. Finally, we formulate the inference-time obstacle-avoidance problem
for a pretrained 3D diffusion policy.

\subsection{Visuomotor Control Setting}
In this subsection, we define the visuomotor control setting and notation utilized
throughout this paper.
We consider closed-loop visuomotor control for a robot manipulator
operating in a workspace with obstacles. At control step $t$, the robot
receives an observation $o_t=(P_t,s_t)$, where
$P_t \in \mathbb{R}^{M \times 3}$ denotes the point-cloud
observation with $M$ points, and $s_t$ denotes the robot proprioceptive
state, including the end-effector pose and gripper state. The
observation history $O_t=(o_{t-T_o+1},\ldots,o_t)$ is formed by stacking
the most recent $T_o$ observations, where $T_o$ is the observation horizon.
Given $O_t$, the policy predicts an action chunk
$A_t=(a_{t,1},\ldots,a_{t,H})$, with
$a_{t,i}\in\mathbb{R}^{d_a}$,
where $i=1,\ldots,H$ indexes future action steps, $H$ is the prediction
horizon, and $a_{t,i}$ is a 7D robot action with $d_a=7$, including a 6D
end-effector command with three positional and three rotational
components, and a scalar gripper opening/closing command.
The positional command can represent either an absolute end-effector
position or a relative displacement. This work adopts the latter.
Following the
receding-horizon execution scheme of Diffusion Policy
\cite{chi2023diffusionpolicy}, only the first $N \leq H$ actions are
executed before replanning, thereby maintaining closed-loop
responsiveness. We denote the executable prefix as
$A_{t,1:N}=(a_{t,1},\ldots,a_{t,N})$. We refer to this closed-loop process 
of observation-conditioned action prediction and receding-horizon execution as visuomotor control.

\subsection{Guided Diffusion Policy and Problem Formulation}
\label{sec:guided-diffusion}
In this subsection, we first review Diffusion Policy (DP) to establish the
conditional action-generation and denoising formulation~\cite{chi2023diffusionpolicy,Ze2024DP3}.
We then introduce the general inference-time guidance update for
static-obstacle avoidance. Finally, we examine the limitations of existing
approaches in visual feedback and propagation of
obstacle responses across the prediction horizon.

Diffusion policies model the conditional action distribution
$\pi_\theta(A_t \mid O_t)$ through an iterative denoising process. Let $A_t^0$ denote a clean action
chunk sampled from the demonstration data distribution. In the forward
diffusion process~\cite{ho2020ddpm}, Gaussian noise is gradually added to
the action chunk:
\begin{equation}
    q(A_t^r \mid A_t^{r-1})
    =
    \mathcal{N}
    \left(
    A_t^r;
    \sqrt{1-\beta_r}A_t^{r-1},
    \beta_r I
    \right),
\end{equation}
where $r=1,\ldots,R$ denotes the forward diffusion step and $\{\beta_r\}_{r=1}^R$
is a predefined noise schedule that controls how much Gaussian noise is
added at each diffusion step. Equivalently, the noisy action chunk at step
$r$ has the following closed-form expression in terms of the clean action
chunk $A_t^0$:
\begin{equation}
    q(A_t^r \mid A_t^0)
    =
    \mathcal{N}
    \left(
    A_t^r;
    \sqrt{\bar{\alpha}_r}A_t^0,
    (1-\bar{\alpha}_r)I
    \right),
\end{equation}
where $\alpha_r=1-\beta_r$ and
$\bar{\alpha}_r=\prod_{j=1}^{r}\alpha_j$.

During training, a clean action chunk is first perturbed by the forward
diffusion process, and the denoising network is trained to predict the
added noise: $\mathcal{L}_{\epsilon}
=\mathrm{MSE}(\epsilon,\epsilon_\theta(A_t^r,O_t,r))$, where $\epsilon$ is
the Gaussian noise utilized to construct $A_t^r$.

During deployment, they may utilize either DDPM~\cite{ho2020ddpm} or
DDIM~\cite{song2021ddim} sampling with $K$ reverse
denoising steps, indexed by $k=K,\ldots,1$. Action generation starts from Gaussian noise
$A_t^K \sim \mathcal{N}(0,I)$ and iteratively denoises it conditioned on
the observation history $O_t$. A denoising network
$\epsilon_\theta(A_t^k, O_t, k)$ predicts the noise component, and the
unguided reverse update can be written as
\begin{equation}
    A_{t,\mathrm{base}}^{k-1}
    =
    \alpha_k'
    \left(
    A_t^k - \gamma_k \epsilon_\theta(A_t^k, O_t, k)
    \right)
    +
    \sigma_k z,
    \quad z \sim \mathcal{N}(0,I),
\end{equation}
where $\alpha_k'$, $\gamma_k$, and $\sigma_k$ are determined by the
chosen sampler, and $A_{t,\mathrm{base}}^{k-1}$ denotes the next sample
produced by the standard reverse process without external guidance.
Under this noise-prediction formulation, the same network prediction
also yields an estimate of the clean action chunk,
$\hat{A}_t^0=(A_t^k-\sqrt{1-\bar{\alpha}_k}\,
\epsilon_\theta(A_t^k,O_t,k))/\sqrt{\bar{\alpha}_k}$,
where $\bar{\alpha}_k$ is the cumulative coefficient at the diffusion
timestep selected for reverse step $k$. The estimate $\hat{A}_t^0$ is
recomputed at every reverse step and its dependence on $k$ is implicit.

To enable inference-time obstacle avoidance, let $\mathcal{B}$ denote the static
environment prior encoding obstacle information in the robot workspace. The
reverse process can then be augmented with guidance derived from $\mathcal{B}$.
This guidance is specified by an objective
$\mathcal{J}(\hat{A}_t^0; O_t, \mathcal{B})$, whose descent direction is
\begin{equation}
    g_k = -\nabla_{\hat{A}_t^0}
    \mathcal{J}(\hat{A}_t^0; O_t, \mathcal{B}).
\end{equation}
The reverse update is then biased along this descent direction:
\begin{equation}
    \label{eq:guided-reverse-update}
    A_t^{k-1}
    =
    A_{t,\mathrm{base}}^{k-1}
    +
    \lambda g_k,
\end{equation}
where $\lambda$ controls the guidance strength. The resulting
$A_t^{k-1}$ serves as the input to the next reverse step.

With the general guidance update established, we now examine how related
obstacle-avoidance methods operate and where their limitations arise, utilizing
our notation for comparison.
MPD~\cite{carvalho2025mpd} is an open-loop joint-space
trajectory planner conditioned on specified initial and goal states rather
than on $O_t$. Without visual feedback, its trajectory
cannot be updated from the evolving task state or execution deviations,
preventing it from serving as a closed-loop task policy that generates $A_t$.
Lan-o3dp~\cite{11127231} predicts absolute end-effector pose targets whose
positional components are denoted by $p_{t,i}$, and computes each point-wise
guidance block independently as
$g_{k,i}^{p,\mathrm{Lan}}
=-\nabla_{p_{t,i}}C_{\mathrm{Lan}}(p_{t,i};\mathcal{B})$
for $i=1,\ldots,H$, where $C_{\mathrm{Lan}}$ denotes its point-wise
obstacle cost.
EmbodiSteer~\cite{wang2026embodisteer} independently solves a local
control-barrier-function quadratic-program (CBF-QP) correction
$\Delta q_i^\star$ for each predicted joint configuration $q_i$ and applies
$q_i\leftarrow q_i+\Delta q_i^\star$.
Both methods apply obstacle corrections separately at each horizon step
without explicitly coupling the resulting responses across the action
sequence. Consequently, these local corrections may not yield a coherent 
action sequence that both avoids obstacles and preserves progress toward the task goal. 
This independent treatment does not explicitly account for cross-step 
dependencies across the prediction horizon.

\textbf{Problem Formulation}
Given the observation history $O_t$, the environment prior $\mathcal{B}$, and a
pretrained diffusion policy $\pi_\theta$ with
fixed denoising network parameters $\theta$, our goal is to bias samples
from the learned task distribution toward action sequences whose
executable prefix $A_{t,1:N}$ avoids collisions between the end-effector
reference point and obstacles and preserves the end-to-end observation-to-action 
task behavior of the pretrained visuomotor policy. In particular, obstacle responses 
should be propagated across the action horizon to coordinate the sampled actions, 
rather than being applied independently at individual steps.

\section{HORIZON-COUPLED OBSTACLE GUIDANCE FOR RELATIVE-ACTION DIFFUSION POLICIES}

\begin{figure*}[!t]
    \centering
    \includegraphics[width=\textwidth]{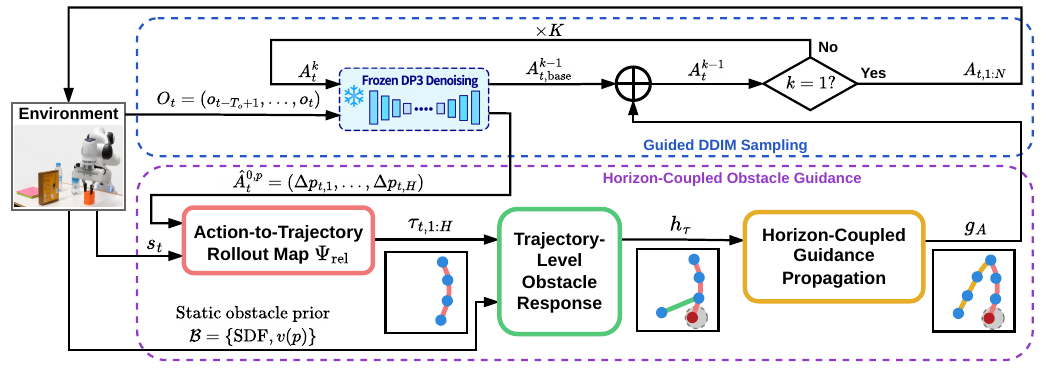}
    \caption{Overview of the proposed framework. Boxes denote processing stages,
    and labeled arrows indicate the variables passed between them. The upper path
    shows guided DDIM sampling, and the lower path shows horizon-coupled obstacle
    guidance. Details are provided in Sec.~\ref{sec:guidance-framework}.}
    \label{fig:overview}
\end{figure*}

In this section, we present inference-time obstacle guidance for a pretrained
3D diffusion policy with relative end-effector position commands.
Our goal is to transform local workspace obstacle responses into
horizon-coupled action updates that coordinate the sampled action sequence
for obstacle avoidance.

\subsection{Horizon-Coupled Obstacle Guidance Framework}
\label{sec:guidance-framework}

We adopt a pretrained DP3 policy~\cite{Ze2024DP3} as the base visuomotor
policy. At each reverse denoising step $k$, given the observation history
$O_t$ and noisy action chunk $A_t^k$, the frozen DP3 policy and DDIM sampler
produce the base reverse update $A_{t,\mathrm{base}}^{k-1}$ and the
corresponding clean action estimate $\hat{A}_t^0$. During guided sampling,
the relative position commands in $\hat{A}_t^0$ are rolled out into a
predicted end-effector trajectory $\tau_{t,1:H}$ and evaluated against the
static environment prior $\mathcal{B}$. The resulting trajectory-level
obstacle responses are propagated across the relative action horizon to
obtain the horizon-coupled guidance $g_A$, which is applied to
$A_{t,\mathrm{base}}^{k-1}$ to produce the guided reverse sample
$A_t^{k-1}$. This process is repeated throughout reverse denoising, with the
pretrained DP3 policy providing the learned task behavior and the proposed
guidance reshaping the sampled action sequence for obstacle avoidance. An
overview of the proposed framework is shown in Fig.~\ref{fig:overview}.

\subsection{Workspace End-Effector Trajectory Evaluation}
\label{sec:trajectory-evaluation}
Obstacle constraints are defined over workspace positions, but the diffusion policy 
predicts relative end-effector position commands rather than workspace trajectories. To bridge this mismatch,
we propose a two-stage workspace trajectory evaluation that rolls out the clean action estimate into a predicted 
end-effector trajectory and evaluates the resulting motion utilizing complementary point-wise and swept-segment obstacle costs.


For the rollout step, we introduce an
action-to-trajectory rollout map $\Psi$, which maps the clean action estimate to a predicted end-effector trajectory in
the workspace given the current robot state $s_t$:
\begin{equation}
    \tau_{t,1:H}
    =
    (p_{t,1}, \ldots, p_{t,H})
    =
    \Psi(\hat{A}_t^0, s_t),
\end{equation}
where $i=1,\ldots,H$ indexes future steps and $p_{t,i}$ denotes the
predicted end-effector workspace position at step $i$. The specific form of $\Psi$ for relative actions
is derived in Sec.~\ref{sec:horizon-propagation}.

Let $C(p;\mathcal{B})$ denote the position-level obstacle cost
evaluated at workspace position $p$ under the static environment prior
$\mathcal{B}$. Its specific design is detailed in
Sec.~\ref{sec:sdf-cost}. To convert workspace obstacle constraints into
a trajectory-level objective for diffusion guidance, we aggregate this
position-level cost along the predicted motion.
The point-wise component of the trajectory-level objective aggregates the
obstacle costs evaluated at the predicted positions along the mapped trajectory
$\tau_{t,1:H}$:
\begin{equation}
    \label{eq:point-obstacle-cost}
    \mathcal{C}_{\mathrm{point}}(\tau_{t,1:H};\mathcal{B})
    =
    \sum_{i=1}^{H} w_i C(p_{t,i};\mathcal{B}),
\end{equation}
where \(w_i\) is an application-dependent horizon weight that modulates the 
relative contribution of the obstacle cost at the \(i\)-th predicted position.

Discrete trajectory points alone may miss collisions between consecutive
predicted positions. For each segment, we therefore sample $M_s$ interior
points as
$\bar{p}_{t,i,m}=(1-\alpha_m)p_{t,i-1}+\alpha_m p_{t,i}$, with
$\alpha_m=m/(M_s+1)$
where $m=1,\ldots,M_s$ indexes the sampled point, $\alpha_m$ specifies
its normalized location along the segment, and $\bar{p}_{t,i,m}$ denotes
the corresponding interpolated workspace position, with $p_{t,0}$
denoting the current end-effector position.  The resulting swept-segment cost
is
\begin{equation}
    \mathcal{C}_{\mathrm{seg}}(\tau_{t,1:H};\mathcal{B})
    =
    \sum_{i=1}^{H}
    \sum_{m=1}^{M_s}
    w_i C(\bar{p}_{t,i,m};\mathcal{B}).
\end{equation}

Combining the point-wise and swept-segment costs, the complete
trajectory-level obstacle objective is
\begin{equation}
    \label{eq:trajectory_obstacle-objective}
    \mathcal{J}_{\mathrm{obs}}(\tau_{t,1:H};\mathcal{B})
    =
    \mathcal{C}_{\mathrm{point}}(\tau_{t,1:H};\mathcal{B})
    +
    \lambda_{\mathrm{seg}}\mathcal{C}_{\mathrm{seg}}(\tau_{t,1:H};\mathcal{B}),
\end{equation}
where $\lambda_{\mathrm{seg}}$ controls the relative strength of the
swept-segment term. Through the rollout map, this trajectory-level objective instantiates the
guidance objective introduced in Sec.~\ref{sec:guided-diffusion} as
$\mathcal{J}(\hat{A}_t^0;O_t,\mathcal{B})
=\mathcal{J}_{\mathrm{obs}}(\Psi(\hat{A}_t^0,s_t);\mathcal{B})$.

\subsection{Horizon-Coupled Propagation for Relative Actions}
\label{sec:horizon-propagation}
In this section, we convert workspace obstacle responses along the predicted 
end-effector trajectory into coordinated updates of the relative position commands. 
We first establish a general Jacobian-based chain-rule mapping, and then derive a 
horizon-coupled propagation rule that propagates obstacle responses across the action horizon, 
from later trajectory points to the preceding increments that contribute to them.

At the end-effector trajectory level, we express the influence of the obstacle
cost in~\eqref{eq:trajectory_obstacle-objective} through its gradient with
respect to the predicted workspace positions.
This trajectory-space gradient is
\vspace{-0.3\baselineskip}
\begin{equation}
    \label{eq:trajectory-gradient}
    h_\tau
    =
    \nabla_{\tau_{t,1:H}}\mathcal{J}_{\mathrm{obs}}(\tau_{t,1:H};\mathcal{B})
    =
    [h_1; \ldots; h_H]
    \in \mathbb{R}^{Hd_p},
\end{equation}
where $h_\tau$ determines the direction and magnitude of the
trajectory-space obstacle response, $d_p$ is the dimensionality of a
workspace position, and each block
$h_i=\partial\mathcal{J}_{\mathrm{obs}}/\partial p_{t,i}
\in\mathbb{R}^{d_p}$ is associated with the $i$-th predicted workspace
position.

To propagate the trajectory-space obstacle response back to the action
variables, we introduce the transpose of the rollout Jacobian:
\begin{equation}
    \label{eq:transposed-rollout-jacobian}
    J_{\Psi,p}(\hat{A}_t^0, s_t)^T
    =
    \left(
    \frac{\partial \Psi(\hat{A}_t^0, s_t)}
    {\partial \hat{A}_t^{0,p}}
    \right)^T
    \in \mathbb{R}^{Hd_p \times Hd_p},
\end{equation}
where $\hat{A}_t^{0,p}=[\hat{a}_{t,1}^{0,p};\ldots;
\hat{a}_{t,H}^{0,p}]\in\mathbb{R}^{Hd_p}$ collects the positional components
of the clean action chunk, with each
$\hat{a}_{t,j}^{0,p}\in\mathbb{R}^{d_p}$.
The transpose of the $(i,j)$ block of $J_{\Psi,p}$ determines how the obstacle response at
the $i$-th predicted workspace position contributes to the positional
component of the $j$-th action.

Combining the trajectory-space obstacle gradient $h_\tau$ with the
transposed rollout Jacobian $J_{\Psi,p}^T$, the chain rule propagates the
obstacle gradient along the reverse variable path
$\mathcal{J}_{\mathrm{obs}}\mapsto\tau_{t,1:H}
\mapsto\hat{A}_t^{0,p}$.
Applying the chain rule through this trajectory-to-action mapping yields the positional guidance:
\begin{equation}
    g_A^p
    =
    -\nabla_{\hat{A}_t^{0,p}}
    \mathcal{J}_{\mathrm{obs}}(\tau_{t,1:H};\mathcal{B})
    =
    -J_{\Psi,p}^T h_\tau ,
\end{equation}
where $g_A^p=[g_1^p;\ldots;g_H^p]$ and $g_j^p$ is the positional
guidance applied to the $j$-th action for $j=1,\ldots,H$.

With the general trajectory-to-action mapping established, we now derive how 
the cumulative relative-action rollout propagates obstacle responses across the action horizon. We adopt this
representation because its cumulative rollout establishes explicit
dependencies across the action chunk. For relative actions,
each positional component is interpreted as an incremental displacement in
the workspace coordinate frame,
$\hat{a}_{t,j}^{0,p}=\Delta p_{t,j}$ with
$\Delta p_{t,j}\in\mathbb{R}^{d_p}$.
The relative rollout map is
\begin{equation}
    \Psi_{\mathrm{rel}}(\hat{A}_t^0,s_t)
    =
    [p_{t,1};\ldots;p_{t,H}],
    \quad
    p_{t,i} = p_{t,0} + \sum_{j=1}^{i} \hat{a}_{t,j}^{0,p},
\end{equation}
where each predicted position $p_{t,i}$ accumulates all positional action
increments from step $1$ through step $i$ and is therefore affected by every
$\hat{a}_{t,j}^{0,p}$ satisfying $j\leq i$.
The Jacobian blocks of $\Psi_{\mathrm{rel}}$ are
\begin{equation}
    \frac{\partial p_{t,i}}{\partial \hat{a}_{t,j}^{0,p}}
    =
    \begin{cases}
    I_{d_p}, & j \leq i,\\
    0, & j > i,
    \end{cases},
\end{equation}
and the corresponding positional component of $g_j$ becomes
\begin{equation}
    g_j^p
    =
    -
    \frac{\partial \mathcal{J}_{\mathrm{obs}}}
    {\partial \hat{a}_{t,j}^{0,p}}
    =
    -
    \sum_{i=j}^{H} h_i,
\end{equation}
where $g_j^p$ accumulates the obstacle responses from all predicted positions
that depend on the $j$-th increment. Because different increments contribute
to overlapping sets of predicted positions, their guidance terms share
obstacle responses and become coupled across the action chunk.
For comparison, Lan-o3dp~\cite{11127231} and
EmbodiSteer~\cite{wang2026embodisteer} apply step-wise obstacle guidance,
which corresponds to a block-diagonal rollout Jacobian and yields the
point-wise form $g_j^p=-h_j$. This contrast shows how the cumulative relative-action rollout 
establishes cross-step dependencies that enable obstacle responses to propagate across the action horizon.
Obstacle guidance is applied only to the positional action components.
We therefore form
$g_j=[g_j^p;\mathbf{0}_{d_a-d_p}]$ and
$g_A=[g_1;\ldots;g_H]$, leaving the rotational and gripper components
unchanged. Recomputed from the current clean action estimate at each
denoising step, $g_A$ serves as the guidance direction $g_k$
in~\eqref{eq:guided-reverse-update}. 
Together, the workspace end-effector trajectory evaluation in
Sec.~\ref{sec:trajectory-evaluation} and
the Jacobian-based horizon-coupled propagation above map workspace obstacle
responses to coordinated updates of the relative position commands, coupling
guidance across the action horizon rather than treating each step
independently.

\subsection{SDF-Based Obstacle Cost}
\label{sec:sdf-cost}
To instantiate the position-level obstacle cost $C(p;\mathcal{B})$, we represent the static environment prior with discretized signed distance
fields (SDFs) over the workspace. We first convert signed distances into an obstacle cost and then
construct the corresponding workspace response for guidance.

We associate each obstacle region $\Omega_\ell$ with an SDF
$\phi_\ell:\mathbb{R}^{d_p}\rightarrow\mathbb{R}$, which returns the signed
distance from a workspace position $p$ to its boundary. We adopt
$\phi_\ell(p)<0$, $\phi_\ell(p)=0$, and $\phi_\ell(p)>0$ for positions
inside $\Omega_\ell$, on $\partial\Omega_\ell$, and outside $\Omega_\ell$,
respectively.
For multiple obstacles, the workspace SDF is
$\phi(p)=\min_{\ell=1,\ldots,L}\phi_\ell(p)$, and
$\ell^*(p)=\arg\min_{\ell=1,\ldots,L}\phi_\ell(p)$ identifies the active
obstacle at $p$, where $\ell$ indexes the obstacles and $L$ is their total
number.

Negative signed distances indicate obstacle penetration. We therefore define
the penetration depth as $d(p)=\operatorname{ReLU}(-\phi(p))$, which is positive and increases with penetration depth inside an obstacle,
and is zero otherwise. The corresponding position-level obstacle cost is
\begin{equation}
    C(p;\mathcal{B})
    =
    d(p)^2
    =
    \left[\operatorname{ReLU}(-\phi(p))\right]^2,
\end{equation}
where the squared form increasingly penalizes deeper penetration and allows
the cost to approach zero smoothly at the obstacle boundary. The
corresponding workspace response is determined by the cost gradient
$h(p)=\nabla_p C(p;\mathcal{B})=-2d(p)\nabla_p\phi(p)$, where $d(p)$
determines the response magnitude according to the penetration depth and
$\nabla_p\phi(p)$ specifies the outward direction, such that the descent
direction $-h(p)$ pushes more deeply penetrating points more strongly toward
free space.

In practice, $\nabla_p\phi(p)$ obtained from a discretized SDF may change
abruptly. We therefore replace it with a precomputed outward direction field
$v(p)$, which serves as a heuristic direction for workspace guidance. For multiple convex
obstacle regions, each with an interior reference center $c_\ell$, the
direction at $p$ is determined by the active obstacle:
\begin{equation}
    v(p)
    =
    \frac{p-c_{\ell^*(p)}}
    {\|p-c_{\ell^*(p)}\|+\epsilon},
\end{equation}
where $c_{\ell^*(p)}$ is the reference center of the active obstacle and
$\epsilon$ is a small constant for numerical stability. 
The resulting heuristic workspace response, distinguished from the exact cost
gradient $h(p)$, is
\begin{equation}
    \tilde{h}(p)=-2d(p)v(p),
\end{equation}
which combines penetration-dependent magnitude with a consistent spatial
direction and allows the descent update $-\tilde{h}(p)$ to push penetrating points
toward free space. The heuristic local responses evaluated at the predicted and interpolated positions
are weighted according to~\eqref{eq:point-obstacle-cost}--\eqref{eq:trajectory-gradient}, with each interpolated response
distributed to its two segment endpoints according to $\alpha_m$ and
$1-\alpha_m$. These contributions form each $\tilde{h}_i$, whose concatenation
gives $\tilde{h}_\tau=[\tilde{h}_1;\ldots;\tilde{h}_H]$. In the implemented
guidance, $\tilde{h}_\tau$ replaces the exact cost gradient $h_\tau$ in the
horizon-coupled propagation above.

Overall, Sec.~\ref{sec:guidance-framework} integrates the proposed guidance
into the diffusion sampling process. Sec.~\ref{sec:trajectory-evaluation}
maps the predicted relative actions to an end-effector trajectory for obstacle
evaluation. Sec.~\ref{sec:horizon-propagation} derives the horizon-coupled
propagation from trajectory-level obstacle responses to action guidance, and
Sec.~\ref{sec:sdf-cost} specifies the SDF-based obstacle cost and practical
workspace response. Algorithm~\ref{alg:sdf_guided_inference} summarizes the
resulting inference procedure.

\begin{algorithm}[t]
\caption{Horizon-Coupled Obstacle Guidance for Diffusion Policy Inference}
\label{alg:sdf_guided_inference}
\begin{algorithmic}[1]
\REQUIRE Observation history $O_t$, environment prior $\mathcal{B}$,
denoising network $\epsilon_\theta$
\STATE $A_t^K\sim\mathcal{N}(0,I)$
\FOR{$k=K,\ldots,1$}
    \STATE $\epsilon_k\leftarrow\epsilon_\theta(A_t^k,O_t,k)$
    \STATE
    $(A_{t,\mathrm{base}}^{k-1},\hat{A}_t^0)$
    $\leftarrow\mathrm{DDIMStep}(A_t^k,\epsilon_k,k)$
    \STATE $\tau_{t,1:H}\leftarrow\Psi_{\mathrm{rel}}(\hat{A}_t^0,s_t)$
    \STATE $\tilde{h}(p)\leftarrow-2d(p)v(p),\quad
    p\in\{p_{t,i},\bar{p}_{t,i,m}\}$
    \STATE $\tilde{h}_\tau\leftarrow[\tilde{h}_i]_{i=1}^{H}$ from weighted
    point and segment responses
    \STATE $g_A\leftarrow[g_j]_{j=1}^{H},\quad
    g_j=[-\sum_{i=j}^{H}\tilde{h}_i;\mathbf{0}_{d_a-d_p}]$
    \STATE $A_t^{k-1}\leftarrow A_{t,\mathrm{base}}^{k-1}+\lambda g_A$
\ENDFOR
\STATE Execute $A_{t,1:N}$ from $A_t^0$ and replan
\end{algorithmic}
\end{algorithm}

\section{Implementation Details}
\label{sec:implementation}

\begin{figure}[t]
    \centering
    \includegraphics[width=0.82\columnwidth]{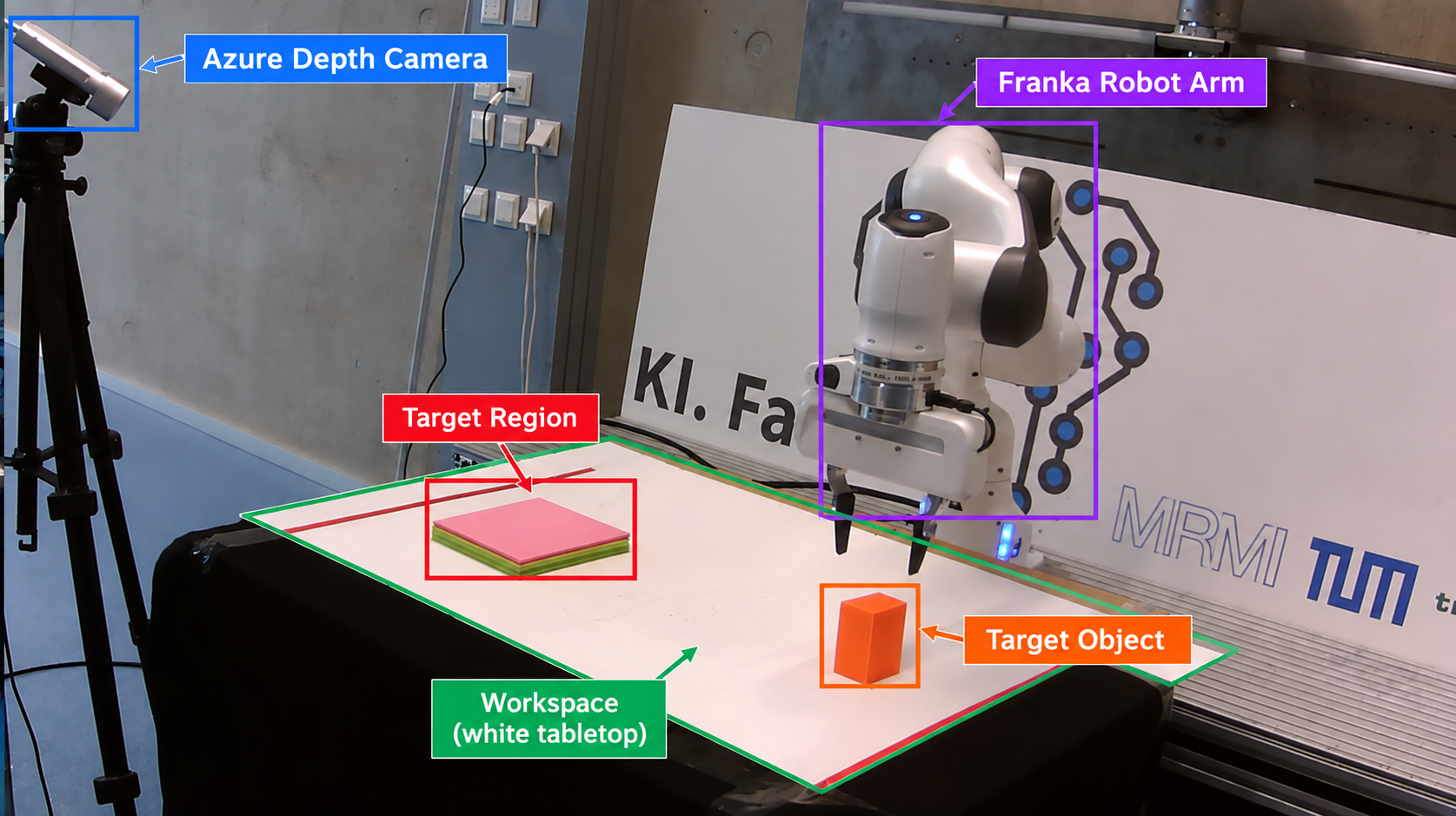}
    \caption{Real-robot experimental system and workspace.}
    \label{fig:system_setup}
\end{figure}

\textbf{System Implementation.}
Data collection and real-robot inference were implemented with the Robot
Control Stack (RCS)~\cite{juelg2026robotcontrolstack}. Demonstrations were
collected through teleoperation with a PlayStation 5 DualSense controller at
20~Hz in obstacle-free workspaces and spanned multiple task-completing
trajectory modes to capture the diversity of valid behaviors. The
demonstrations were downsampled from 20~Hz to 1~Hz before policy training to
match the deployment action frequency.
During deployment, the robot executed incremental actions at 1~Hz, which is a
reasonable control frequency for these manipulation tasks.

\textbf{Policy Configuration.}
We set the observation horizon to $T_o=2$. Within each observation
$o_t$, the end-effector orientation in the proprioceptive state $s_t$ was
encoded in the continuous 6D rotation
representation~\cite{geist2024rotations}, which avoids angular wrap-around
and quaternion double-cover ambiguity and provides a smooth Euclidean input
to the policy. This yielded $s_t\in\mathbb{R}^{10}$, consisting of the
end-effector position, 6D rotation, and gripper width. Point clouds were
cropped to the tabletop workspace and uniformly sampled to 1024 points.
The policy predicted action chunks over a horizon of $H=16$ and executed the first $N=8$
actions of each sampled chunk. The denoising network was configured for
direct sample prediction, with $R=100$ forward diffusion steps during training
and $K=10$ DDIM reverse steps during inference. The policy was trained for 101 epochs with a
batch size of 128. We selected DDIM because it permits fewer
denoising steps during inference, thereby reducing inference
latency~\cite{chi2023diffusionpolicy,song2021ddim}.

\textbf{Guidance Implementation.}
Each trajectory segment contained
$M_s=10$ samples, with the segment weight set to
$\lambda_{\mathrm{seg}}=10$. The guidance strength $\lambda$ was selected
empirically for each experimental scenario to balance obstacle avoidance and
task execution. At each DDIM step, the clean action estimate was transformed 
back to the original action space for workspace guidance, and the resulting guidance update was rescaled before being added to the normalized reverse sample.

\section{Experimental Evaluation}
\noindent To examine the benefits of the proposed method, we conduct a series of
real-robot experiments addressing four questions:
\textbf{Q1:} Can our method improve collision-free task success?
\textbf{Q2:} How does our method compare with the baseline in obstacle avoidance?
\textbf{Q3:} How do obstacle modeling and guidance formulation affect obstacle avoidance and task success?
\textbf{Q4:} Can the same guidance formulation support diverse manipulation tasks?

\begin{figure*}[t]
    \centering
    \setlength{\tabcolsep}{1.2pt}
    \begin{tabular}{@{}cccc@{}}
        \includegraphics[width=0.244\textwidth]{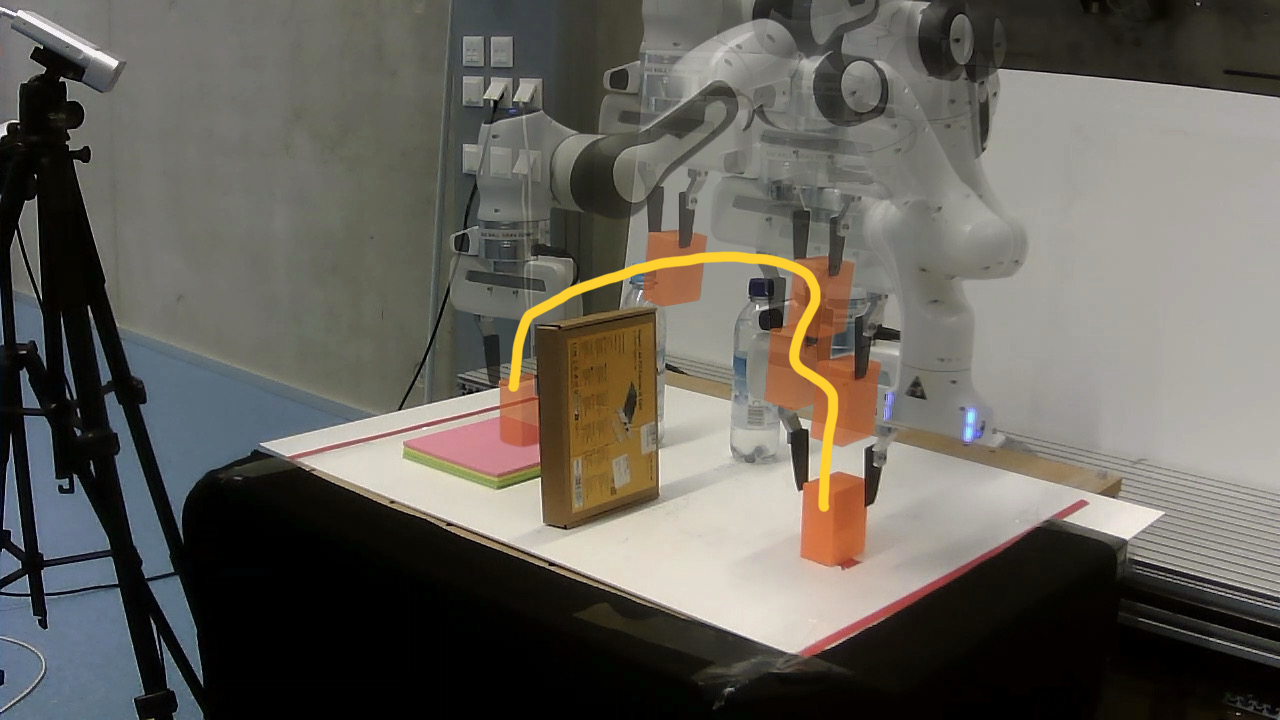} &
        \includegraphics[width=0.244\textwidth]{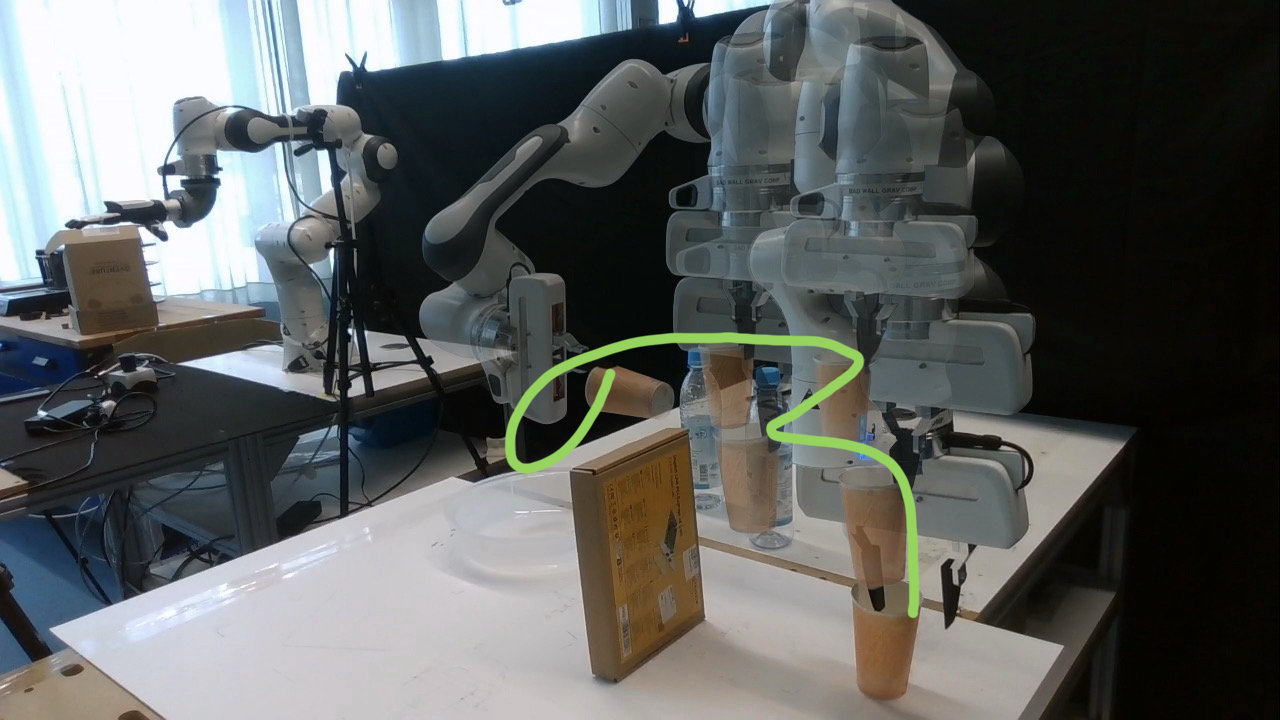} &
        \includegraphics[width=0.244\textwidth]{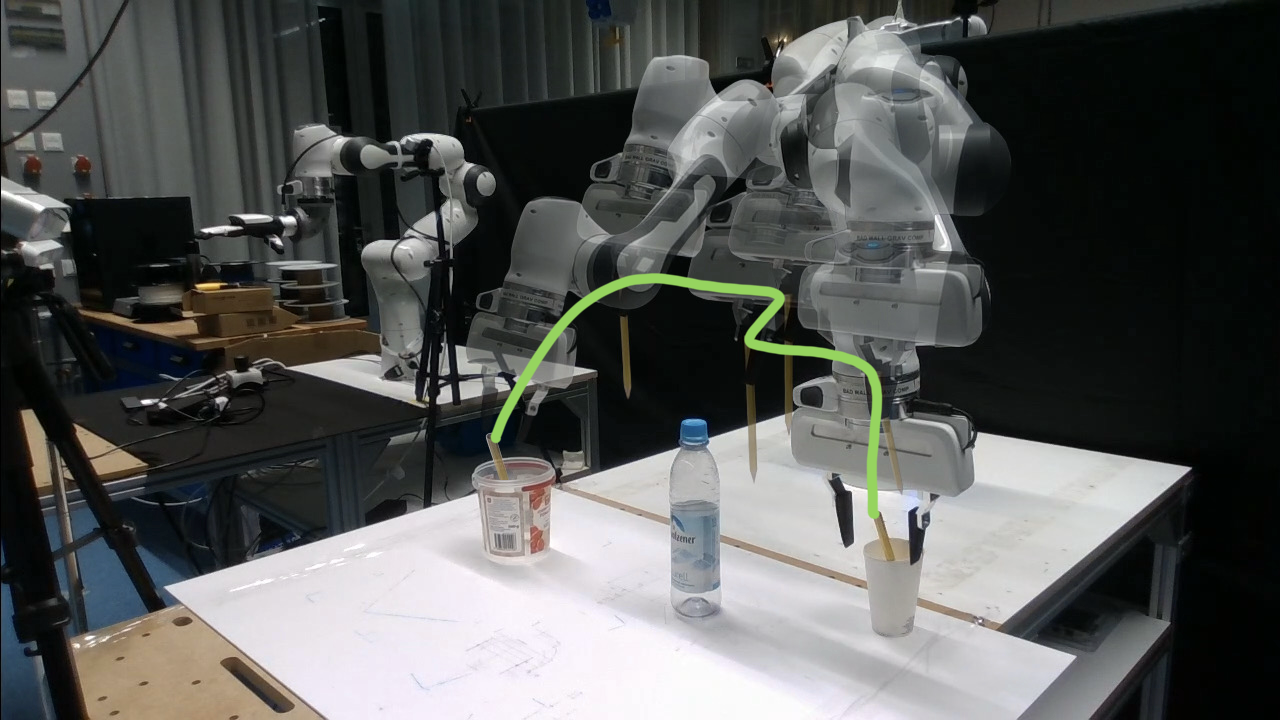} &
        \includegraphics[width=0.244\textwidth]{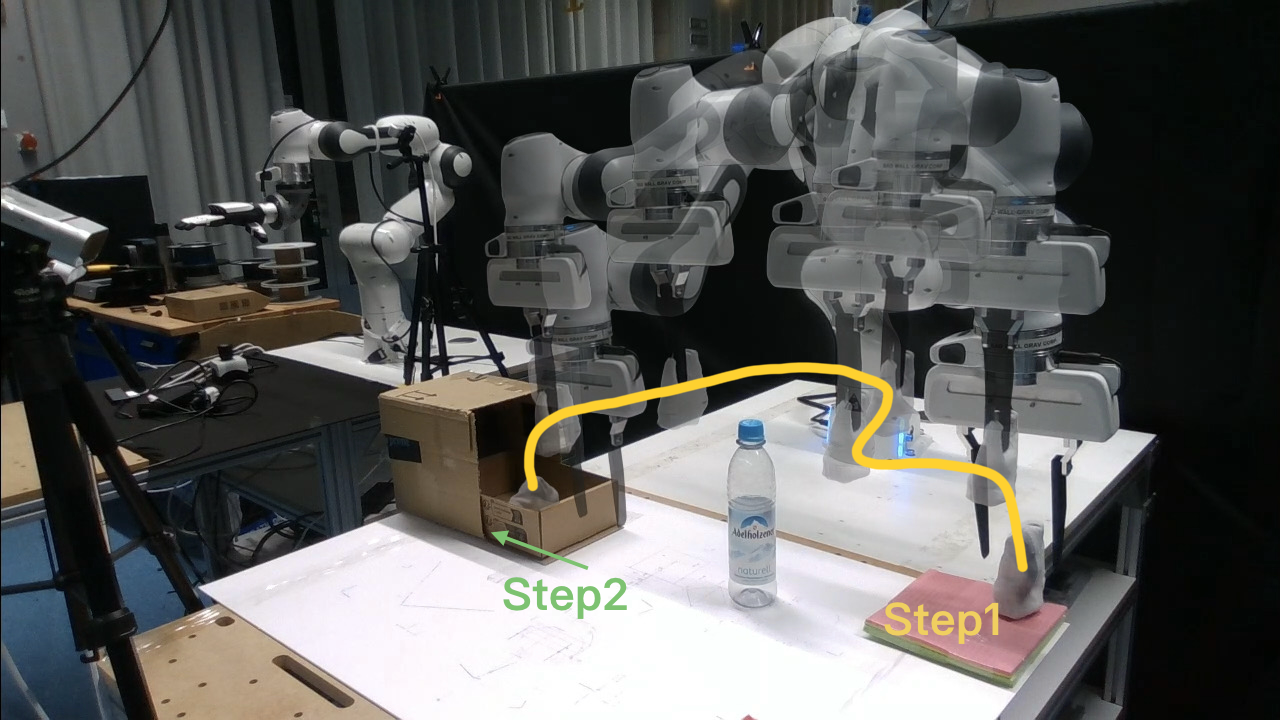} \\[-1pt]
        \small\textbf{(a) Pick-and-place} &
        \small\textbf{(b) Pouring} &
        \small\textbf{(c) Pen insertion} &
        \small\textbf{(d) Sock + drawer}
    \end{tabular}
    \caption{Our four real-world manipulation tasks span obstacle-constrained
    pick-and-place, pouring with end-effector roll, pen insertion with pitch
    adjustment, and sequential sock placement and drawer closing. Superimposed
    robot poses visualize each execution sequence.}
    \label{fig:real_robot_executions}
\end{figure*}

\subsection{Experimental Setup}

\subsubsection{Platform and Manipulation Tasks}
Experiments utilized a 7-DoF Franka Emika Panda robot arm and an Azure Kinect
DK depth camera providing third-person-view point clouds, with 3D-printed
finger extensions improving gripper visibility.
Data collection and inference were performed on an RTX 3050 Ti laptop, and
policy training on an RTX 4060 laptop. On the deployment computer, the policy achieved
inference rates of 4.10~Hz without guidance and 2.16~Hz with obstacle
guidance. The experimental system is shown in Fig.~\ref{fig:system_setup}.

We evaluate four real-robot manipulation tasks with distinct motion
requirements. \textsc{Pick-and-Place} requires transporting a cube to a target
region on the opposite side of the table. \textsc{Pouring} additionally
requires an end-effector roll motion, and \textsc{Pen Insertion} requires
pitch adjustment followed by insertion. \textsc{Sock + Drawer} is a two-stage
task in which the robot grasps a sock, transports it into an open drawer, and
then closes the drawer. Representative executions of all four tasks are shown
in Fig.~\ref{fig:real_robot_executions}.

\subsubsection{Evaluation Scenarios}
Within each scenario, the robot initial configuration, object arrangement,
and task target were fixed across trials. We first evaluate all four tasks
without obstacles to establish the nominal task performance of the pretrained
policies. For \textsc{Pick-and-Place}, we evaluate four static-obstacle
layouts: a single rectangular obstacle, a two-obstacle narrow passage, and
three- and four-obstacle layouts with progressively constrained free space.
The single-obstacle layout is also utilized for the component ablation.
For cross-task evaluation, static obstacles are placed along the transport
region of \textsc{Pouring}, \textsc{Pen Insertion}, and
\textsc{Sock + Drawer}.

\subsubsection{Baselines, Training and Ablation Design}
To evaluate the effectiveness of our method, we construct four evaluation
variants. For each manipulation task, we collect 50 obstacle-free expert
demonstrations. We adopt DP3~\cite{Ze2024DP3} as the policy backbone and 
independently train one policy with absolute end-effector pose actions and 
one with relative end-effector pose actions under the same demonstrations, 
training configuration, and number of optimization steps.
The training hyperparameters are specified in Section~\ref{sec:implementation}.

Without obstacle guidance, the pretrained absolute- and relative-action 
policies are denoted as \textsc{Abs.-NoG} and \textsc{Rel.-NoG}, respectively.
To represent the step-wise obstacle-guidance methods~\cite{11127231,wang2026embodisteer}
discussed in Section~\ref{sec:preliminaries}, which apply obstacle corrections independently at
individual horizon steps, we define \textsc{Abs.-Guide} by augmenting the same
absolute-action policy with Lan-o3dp's point-wise guidance~\cite{11127231}
and its circular or elliptical primitive envelopes. \textsc{Abs.-Guide} shares 
the same pretrained absolute-action policy as \textsc{Abs.-NoG}. \textsc{Ours} applies horizon-coupled
obstacle guidance to the relative-action policy with SDF-based obstacle
modeling. \textsc{Rel.-NoG} and \textsc{Ours} share the same pretrained relative-action
policy.

The ablation first compares \textsc{Abs.-NoG} and \textsc{Rel.-NoG} to
test whether relative actions alone improve obstacle avoidance. We then utilize
a $2\times2$ design combining primitive-envelope and SDF obstacle models
with absolute-action point-wise and relative-action horizon-coupled guidance.
All obstacle-model variants within the same guidance formulation share the
same pretrained policy.

\subsubsection{Evaluation Protocol}
All policy variants were evaluated under identical physical obstacle
placements within each scenario. Each obstacle's geometry and workspace
coordinates were measured manually and kept fixed across repeated trials to
construct the environment prior $\mathcal{B}$ described in
Sec.~\ref{sec:sdf-cost}. 
Obstacles were evaluated in the horizontal $x$--$y$ plane, and guidance
was applied only to the corresponding positional action components. The
guidance strength $\lambda$ was selected empirically for each guided
configuration and experimental scenario to balance obstacle avoidance and
task execution.

Each method--scenario combination was evaluated over 20 real-robot trials.
A trial terminated upon task completion or failure. Failures were 
categorized as collisions or non-collision execution failures. Non-collision execution 
failures included failed grasps, object drops, failure to reach the target region, or inability to continue the task.
Task-specific completion was defined as follows: \textsc{Pick-and-Place}
required the cube to be placed on the marked mat; \textsc{Pouring} required
completion of the pouring motion with the paper cup above the circular target
pad; \textsc{Pen Insertion} required the pen to be inserted into the
cylindrical pen cup; and \textsc{Sock + Drawer} required the sock to be placed
inside the drawer before the drawer was closed. A collision was recorded when
the robot or manipulated object made any observable contact with an obstacle.
A collision or execution failure was counted as a failed trial. Collision
outcomes were determined through direct observation and video review.
A trial was counted as a collision-free task success only when its completion criterion
was satisfied without obstacle contact. 
\subsection{Real-Robot Results and Analysis}

\begin{table*}[t]
    \caption{Collision-free task successes across (a) obstacle configurations
    and (b) manipulation tasks.}
    \label{tab:obstacle_results}
    \centering
    \footnotesize
    \renewcommand{\arraystretch}{1.12}
    \setlength{\tabcolsep}{2pt}
    \begin{tabular*}{\textwidth}{@{\extracolsep{\fill}}l|ccccc|cccc|c@{\hspace{2em}}}
        \hline
        \multicolumn{1}{c|}{\multirow{2}{*}[-0.5ex]{Method}}
        & \multicolumn{5}{c|}{\textbf{(a) Obstacle-Configuration Evaluation}}
        & \multicolumn{4}{c|}{\textbf{(b) Cross-Task Evaluation}}
        & \multirow{2}{*}[-0.5ex]{All 7} \\
        \cline{2-10}
        & Rect. & 2 Obs. & 3 Obs. & 4 Obs. & Overall
        & Pouring & \shortstack{Pen\\Insertion}
        & \shortstack{Sock +\\Drawer} & Overall & \\
        \hline
        Abs.-NoG
        & 2/20 & 0/20 & 0/20 & 0/20 & 2/80
        & 0/20 & 0/20 & 2/20 & 2/60 & 4/140 \\
        Rel.-NoG
        & 0/20 & 0/20 & 0/20 & 0/20 & 0/80
        & 0/20 & 0/20 & 0/20 & 0/60 & 0/140 \\
        Abs.-Guide
        & 2/20 & 1/20 & 1/20 & 0/20 & 4/80
        & 0/20 & 0/20 & 4/20 & 4/60 & 8/140 \\
        \cmidrule{1-11}
        \textbf{Ours}
        & \textbf{18/20} & \textbf{16/20}
        & \textbf{12/20} & \textbf{8/20} & \textbf{54/80}
        & \textbf{20/20} & \textbf{16/20}
        & \textbf{15/20} & \textbf{51/60} & \textbf{105/140} \\
        \cmidrule{1-11}
    \end{tabular*}
\end{table*}

\begin{table}[t]
    \caption{Task successes in obstacle-free reference settings.}
    \label{tab:obstacle_free_results}
    \centering
    \footnotesize
    \renewcommand{\arraystretch}{1.12}
    \begin{tabular*}{\columnwidth}{@{\extracolsep{\fill}}l|cccc@{}}
        \toprule
        Method & \shortstack{Pick-and-\\Place} & Pouring
        & \shortstack{Pen\\Insertion} & \shortstack{Sock +\\Drawer} \\
        \midrule
        Abs.-NoG & 20/20 & 20/20 & 13/20 & 20/20 \\
        Rel.-NoG & 20/20 & 20/20 & 18/20 & 19/20 \\
        \bottomrule
    \end{tabular*}
\end{table}

\noindent\textbf{1) Collision-Free Task Success:}
Tables~\ref{tab:obstacle_results}(a) and~\ref{tab:obstacle_free_results}
jointly answer Q1. Table~\ref{tab:obstacle_free_results} confirms the nominal
task competence of the pretrained absolute- and relative-action policies,
which both achieve 20/20 success in obstacle-free pick-and-place. Across the
four obstacle layouts, \textsc{Abs.-NoG} achieves only 2/80 collision-free task
successes, and \textsc{Rel.-NoG} achieves 0/80. Thus, unguided execution of
policies trained without obstacles does not reliably satisfy deployment-time
obstacle constraints.
\textsc{Abs.-Guide} improves collision-free task success from 2/80 to 4/80
compared with \textsc{Abs.-NoG}. \textsc{Ours} achieves 54/80 compared with
0/80 for \textsc{Rel.-NoG} and is the only method with nonzero success in every
layout, achieving 18/20, 16/20, 12/20, and 8/20 successes. These results answer Q1 affirmatively: our
method substantially improves collision-free task success over both unguided
execution and point-wise obstacle guidance across all tested obstacle layouts.

\begin{figure}[t]
    \centering
    \includegraphics[width=\columnwidth]{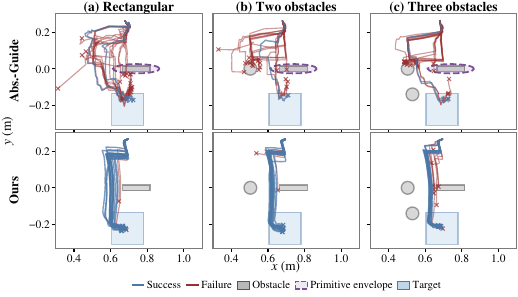}
    \caption{Executed end-effector trajectories of \textsc{Abs.-Guide} (top)
    and \textsc{Ours} (bottom) across rectangular-, two-, and three-obstacle
    layouts. Purple dashed contours denote the primitive envelopes utilized by
    \textsc{Abs.-Guide}. Blue and red trajectories indicate successful and failed
    trials, and crosses mark terminal positions.}
    \label{fig:trajectory_comparison}
\end{figure}

\noindent\textbf{2) Obstacle-Avoidance Behavior:}
To answer Q2, Fig.~\ref{fig:trajectory_comparison} shows how the performance
gap in Q1 manifests in the end-effector trajectories under progressively
restricted free space. In Fig.~\ref{fig:trajectory_comparison}(a),
our trajectories follow a consistent detour around the rectangular obstacle
before continuing toward the target. The \textsc{Abs.-Guide} paths vary
substantially, and many terminate near or beyond the obstacle without reaching
the target region.
Adding the circular obstacle in Fig.~\ref{fig:trajectory_comparison}(b)
restricts the wider detour to a narrow passage between the obstacles. Our trajectories
follow this passage and continue toward the target. Many \textsc{Abs.-Guide}
executions terminate before or within the passage; others pass between the
obstacles but stop outside the target region. The third obstacle in
Fig.~\ref{fig:trajectory_comparison}(c) further narrows the free-space corridor
in the later part of the motion. Our trajectories remain largely confined to
this corridor and extend toward the target region. The \textsc{Abs.-Guide}
trajectories take different routes and do not consistently follow the
remaining free-space corridor.
Across the three layouts, our method more often forms a complete avoidance
route that follows the available free space and remains directed toward the
task goal. 
\begin{figure}[t]
    \centering
    \includegraphics[width=\columnwidth]{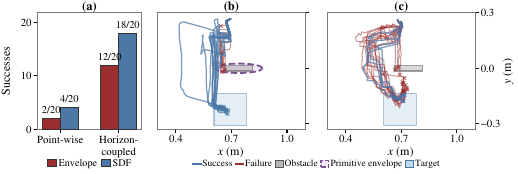}
    \caption{Ablation of obstacle modeling and guidance formulation.
    (a) Collision-free task successes for the four combinations.
    Executed end-effector trajectories under (b) primitive-envelope
    relative-action horizon-coupled guidance and (c) SDF absolute-action
    point-wise guidance. Purple dashed contours denote the primitive envelope.}
    \label{fig:ablation}
\end{figure}

\noindent\textbf{3) Component Ablation:}
To answer Q3, we first determine whether relative actions alone improve
obstacle avoidance. Table~\ref{tab:obstacle_free_results} shows that both
unguided policies complete all obstacle-free \textsc{Pick-and-Place} trials.
In the single-rectangular-obstacle setting of
Table~\ref{tab:obstacle_results}(a), \textsc{Rel.-NoG} does not improve
collision-free task success over \textsc{Abs.-NoG}. Relative actions alone
therefore cannot account for the improvement of our method. We then utilize Fig.~\ref{fig:ablation}(a) to examine the effects of obstacle
modeling and guidance formulation. Relative-action horizon-coupled guidance
outperforms absolute-action point-wise guidance by 10 successes with the
primitive envelope and 14 with the SDF. SDF-based modeling improves success
by 2 under point-wise guidance and 6 under horizon-coupled guidance. The
relative-action horizon-coupled formulation therefore provides the larger
observed gain, with SDF-based obstacle modeling contributing a complementary
improvement.

The trajectory comparisons reveal how the two factors affect executed
motion. With the primitive envelope, absolute-action point-wise guidance in the upper panel 
of Fig.~\ref{fig:trajectory_comparison}(a) frequently terminates before reaching the target. 
Fig.~\ref{fig:ablation}(b) shows that relative-action horizon-coupled guidance more often forms 
complete detours around the obstacle and back toward the target. The same pattern appears with the SDF:
absolute-action point-wise trajectories in Fig.~\ref{fig:ablation}(c) remain
mostly unsuccessful, whereas the horizon-coupled trajectories in the lower
panel of Fig.~\ref{fig:trajectory_comparison}(a) form a concentrated family
of task-directed detours. These results are consistent with the horizon-coupled propagation derived in
Sec.~\ref{sec:horizon-propagation}. An obstacle response at a later predicted
position is propagated
backward to the preceding relative position increments that contribute to
that position, coordinating the sampled actions across the horizon.

Holding this guidance formulation fixed, comparing Fig.~\ref{fig:ablation}(b) with
the lower panel of Fig.~\ref{fig:trajectory_comparison}(a) isolates the effect
of obstacle modeling. Primitive-envelope trajectories are more dispersed and
include occasional large lateral deviations, whereas SDF trajectories more
closely follow a common route around the obstacle. These results indicate that
horizon-coupled propagation primarily enables coordinated, task-directed
avoidance motion, and SDF-based modeling further shapes the detour according
to the obstacle geometry.

\noindent\textbf{4) Cross-Task Applicability:}
Tables~\ref{tab:obstacle_results}(b) and
\ref{tab:obstacle_free_results} jointly answer Q4. The pretrained
relative-action policies achieve 20/20, 18/20, and 19/20 task successes in
the obstacle-free \textsc{Pouring}, \textsc{Pen Insertion}, and
\textsc{Sock + Drawer} settings. With obstacles and the same horizon-coupled obstacle-guidance formulation
, our method achieves 20/20, 16/20, and 15/20 collision-free task
successes. The guidance modifies only the positional action variables, yet
the guided policies preserve the task-specific rotational, gripper, and sequential behaviors
required by the three tasks. These results demonstrate
that the same obstacle-guidance formulation can be applied to independently trained policies with distinct manipulation requirements, without obstacle-specific
demonstrations or policy retraining.


\section{Conclusion and Future Work}
In this work, we introduce horizon-coupled obstacle guidance for pretrained relative-action diffusion policies.
By propagating workspace obstacle responses across the action horizon, our method coordinates the sampled actions for
complete, task-directed avoidance motion without policy retraining.
Real-robot experiments show substantial improvements over relative actions
alone and absolute-action point-wise guidance. SDF-based obstacle modeling
provides a complementary gain, and the same horizon-coupled guidance
formulation transfers across multiple manipulation tasks. Future work will
extend the framework toward open-world deployment by incorporating broader
scene understanding.

\bibliographystyle{IEEEtran}
\bibliography{references}

\end{document}